# Gated recurrent units and temporal convolutional network for multilabel classification

Loris Nanni[1], Alessandra Lumini[2], Alessandro Manfè[1], Riccardo Rampon[1], Sheryl Brahnam[3] and Giorgio Venturin[1]

[1]University of Padova, Via Gradenigo 6, Padova, Italy  loris.nanni@unipd.it; alessandro.manfe@studenti.unipd.it riccardo.rampon@studenti.unipd.it giorgio.venturini@studenti.unipd.it

[2]DISI, University of Bologna, via dell'Università 50, Cesena, Italy, alessandra.lumini@unibo.it

[3] Missouri State University, USA sbrahnam@missouristate.edu

**Abstract**
Multilabel learning tackles the problem of associating a sample with multiple class labels. This work proposes a new ensemble method for managing multilabel classification: the core of the proposed approach combines a set of gated recurrent units and temporal convolutional neural networks trained with variants of the Adam optimization approach. Multiple Adam variants, including novel one proposed here, are compared and tested; these variants are based on the difference between present and past gradients, with step size adjusted for each parameter. The proposed neural network approach is also combined with Incorporating Multiple Clustering Centers (IMCC), which further boosts classification performance.
Moreover, we report an ablation study for assessing the performance improvement that each module of our ensemble allows to obtain.
Multiple experiments on twelve data sets representing a wide variety of multilabel tasks demonstrate the robustness of our best ensemble, which is shown to outperform the state-of-the-art. The MATLAB code for generating the best ensembles in the experimental section will be available at https://github.com/LorisNanni.

## 1 Introduction

Multilabel learning deals with the common problem of associating a sample with multiple labels and has been successfully applied in various domains [1] such as tag recommendation [2], bioinformatics [3-6], information retrieval [7, 8], speech recognition [9, 10], and user reviews and negative comment classification [11, 12], to list a few.

The most intuitive way for handling the multilabel problem is to decompose it into multiple independent binary classification problems, a solution known as Binary Relevance (BR) [13, 14]. Despite its simplicity, BR has several drawbacks: relatively poor performance, lack of scalability, and an inability to retain label correlations. These problems can be addressed with chains of binary classifiers [15], feature selection methods [16-18], class-dependent and label-specific features [19], and data augmentation methods [7]. Of note is the augmentation method presented in [7] that has achieved top performance on some multilabel data sets by incorporating multiple cluster centers (IMCC). For biomedical data sets, one of the best multilabel classifiers is hML-KNN [20], which determines classes by considering both feature-based and neighbor-based similarity scores.

A dependable and robust method for enhancing the performance of multilabel models is to generate ensembles of classifiers [17, 21], with bagging one of the most popular method for producing ensembles that perform well across several multilabel classification benchmarks [22, 23]. In [24], the authors attempt to overcome the problem of pairwise label correlation common in bagging by using a stacking technique where binary classifiers are trained on each label with their predictions used as inputs into stacked combinations via a meta-level classifier that outputs a final decision. Also popular is the Random k-labELsets (RAKEL) algorithm [21], which builds ensembles by training single-label learning models on small random subsets of labels. Some recent variants of RAKEL have been proposed in [25] and [26].

As in the case with other machine learning approaches, deep learning has taken center stage in contemporary approaches for handling multilabel learning. The ascendancy of deep learning in this field is evident in the proliferation of open source and commercial APIs that address deep multilabel classification. Commercial APIs include Imagga [27], Clarifai [28], Microsoft's Computer Vision [29], IBM Watson's Visual Recognition [30], Google's Cloud Vision [31], and Wolfram Alpha's Image Identification [32]. Opensource APIs include DeepDetect [33], VGG19 (VGG) [34], Inception v3 [35], InceptionResNet v2 [36], ResNet50 [37], MobileNet v2 [38], YOLO v3 [39], and fastai [40] (for a comparison of the performance of all but the last API across several multilabel data sets, see [41]).

In this paper, we propose an ensemble for multilabel classification that combines Gated Recurrent Units (GRU) [42] and Temporal Convolutional Neural Networks (TCN) [43]. A GRU can be conceptualized as a simplified Bidirectional Long Short-Term Memory (LSTM), both of which model hidden temporal states with some gating mechanisms and suffer from intermediate activations that are a function of low-level features. For this reason, GRU and LSTM must be combined with other classifiers that could capture high-level relationships. In this work, we investigate TCN as a potential complement as it is designed to hierarchically capture relationships across high, intermediate, and low-level timescales.

To ensure diversity in ensembles, TCN and GRU networks are trained on different variants of Adam [44] optimization. Adam is best known for realizing low minima of the training loss. Ensembles of networks are built by applying different optimization approaches that boost the performance of the original Adam methods. Because Adam variants are unstable, they can augment diversity among classifiers. An additional improvement of performance is obtained by combining the proposed ensemble of GRU and TCN with IMCC [7].

This study makes several contributions. To the best of our knowledge, this is the first work to propose an ensemble method for managing multilabel classification that is based on combining sets of GRU and TCN, and the first to use TCN.

We proposed two different topologies based on GRU and two topologies based on TCN. Moreover, we proposed a topology based on GRU and TCN combined in a sequential way.

In addition, the classifiers are trained on different variants of Adam optimization, including one novel variant proposed here.

The effectiveness of the proposed ensemble is demonstrated by comparing its performance with several baseline approaches across twelve multilabel benchmarks. Experiments show that the proposed method gains state-of-the-art performance across many applications without requiring any tuning. So that the system proposed here can be compared with future works, all source code developed for this study is available at https://github.com/LorisNanni.

The remainder of this paper is organized as follows. In section 2, recent work related to multilabel classification using deep learners and GRU is covered, as well as a description of the twelve data sets and the performance indicators used for evaluation. In section 3, the proposed approach is described along with each element and process involved: preprocessing methods, GRU,

TCN, IMCC, pooling, training, and ensemble creation. In section 4, Adam optimization and the novel Adam variants proposed here are detailed. In section 5, experimental results are presented and discussed. Finally, in section 6, we conclude by summarizing the results and suggesting future directions of research.

## 2   Related works

An early application of deep learning to the problem of multilabel classification was reported in [45], where a simple feed-forward network was implemented to handle the functional genomics problem in computational biology. Within the last few years, there has been a growing body of work applying deep learning to a host of multilabel problems: land cover scene categorization [46] using a Convolutional Neural Network (CNN) combined with data augmentation; detection of heart rhythm/conduction abnormalities [47] using a CNN where each element in the output vector corresponds to a specific rhythm class; multilabel classification of protein-lncRNA interactions [48] using a CNN and an LSTM; Anatomical Therapeutic Chemical (ATC) multilabel classification [3] using an ensemble combining an LSTM Network with a set of classifiers based on Multiple Linear Regression (MLR); surgical tool detection in laparoscopic videos [49] and a recommendation system for pre-diagnosis support [50] using Recurrent CNNs (RCNNs). Other papers based Recurrent CNNs for multilabel classification are [72,73].

In the last couple of years, a growing body of literature has explored the benefits of adding GRUs in multilabel systems. In [51], for example, the authors analyzed sentiment in tweets by taking the topics extracted by a C-GRU (Context-aware GRU). In [52], a new computational method called NCBRPred was proposed to predict the nucleic acid binding residues based on the multilabel sequence labeling model, which used bidirectional GRUs (BiGRUs) to capture the global interactions among the residues. Finally, in [53], the authors proposed a system composed of an Inception model and GRU to identify nine classes of arrhythmias.

## 2.1 Data Sets

Evaluation of our proposed approach requires data sets with binary multilabel categorization. The following twelve data sets were selected because they represent a wide range of applications (such as image, music, biology, and drug classification) and because they are often used when comparing multilabel classification systems (note: the names of the data sets are those typically used in the literature and not necessarily those in the original papers):

1. Cal500 [54]: this is a collection of human-generated annotations that describe popular Western music tracks produced by 500 unique artists. Cal500 includes 502 instances represented by 68 numeric features and 174 distinct labels.

2. Scene [13]: this is a collection of 2407 color images (divided into training and testing images) of different scenes grouped into six base categories *beach* (369), *sunset* (364), *fall foliage* (360), *field* (327), *mountain* (223), and *urban* (210), with sixty-three images belonging to two categories and only one to three categories. Taking into account the joint categories, the total number of labels is fifteen. Each image in this data set went through a four-step preprocessing procedure. In step 1, an image was converted to the CIE Luv space since this space is perceptually uniform (i.e., close to Euclidean distances). In step 2, the image was divided into 49 blocks with a 7×7 grid. In step 3, the mean and variance of each band were computed. The mean is equivalent to a low-resolution image, whereas the variance corresponds to computationally inexpensive texture features. Finally, in step 4, the image was transformed into a 294-dimensional feature vector (49×3×2).

3. Image [55]: this is a collection of 2,000 natural scene images grouped into five base categories *desert* (340), *mountains* (268), *sea* (341), *sunset* (216), and *trees* (378) that intended to produce a much larger set of images than scene that belong to two categories (442) and three categories (15). Taking into account the joint categories, the total number of labels is 20. The images in this data set went through the same preprocessing procedure as in [13].

4. Yeast [2]: this is a biologic data set for classifying 2417 micro-array expression data and phylogenetic profiles (represented by 103 features) into 14 functional classes. A gene can belong to more than one class.

5. Arts [7]: this data set was built using 5000 art images, each described by 462 numeric features where each image can belong to some 26 classes.

6. Liu [17]: this is a data set of drugs collected to predict drug side effects. It includes 832 compounds represented by 2892 features and 1385 labels.

7. ATC [56]: this is a collection of 3883 ATC coded pharmaceuticals with each sample represented by 42 features and 14 classes.

8. ATC_f: this is a variant of the ATC collection where the same patterns are represented by an 806-dimensional descriptor (i.e., all the three descriptors are tested as in [3]).

9. mAn [6]: this is a data set of proteins represented by 20 features and 20 labels.

10. Bibtex, Enron, Health: these are three highly sparse datasets used in [7]

A summary of all the data sets, including the number of patterns, features, labels, and the average number of class labels per pattern (LCard), is reported in Table 1. A 5-fold cross-validation testing protocol with results averaged is used for data set 6, and a 10-fold protocol for data sets 7-9. Data sets 1-5 are in the MATLAB IMCC toolkit [7] available at https://github.com/keauneuh/Incorporating-Multiple-Cluster-Centers-for-Multi-Label-Learning/tree/master/IMCCdata (accessed 9/9/21). All other data sets can be obtained from the authors in the references provided above.

| Name | #patterns | #features | #labels | LCard |
|---|---|---|---|---|
| CAL500 | 502 | 68 | 174 | 26.044 |
| Image | 2000 | 294 | 5 | 1.236 |
| Scene | 2407 | 294 | 5 | 1.074 |
| Yeast | 2417 | 103 | 14 | 4.24 |
| Arts | 5000 | 462 | 26 | 1.636 |
| ATC | 3883 | 42 | 14 | 1.265 |
| ATC_f | 3883 | 700 | 14 | 1.265 |
| Liu | 832 | 2892 | 1385 | 71.160 |
| mAn | 3916 | 20 | 20 | 1.650 |
| bibtex | 7395 | 1836 | 159 | 2.402 |
| enron | 1702 | 1001 | 53 | 3.378 |
| health | 5000 | 612 | 32 | 1.662 |

**Table 1.** Summary of the twelve data sets tested in this work.

### 2.2 Performance indicators

Multiclass classification is evaluated using several performance indicators so that results can be compared with previous studies. Let $X$ be a data set that includes $m$ samples $x_i \in \Re^d$ with each sample having an actual label $y_i \in \{0, 1\}^l$, where $l$ is the number of labels. Letting $H$ and $F$ be the set of predicted labels, where $h_i \in \{0, 1\}^l$ is the predicted label vector for sample $x_i$, and $f_i \in \Re^l$ be the confidence relevance of each prediction, the following performance indicators [7] can be defined for $H$ and $F$:

- Hamming loss: this is the fraction of misclassified labels,

$$\text{HLoss}(H) = \frac{1}{ml}\sum_{i=1}^{m}\sum_{j=1}^{l} I(\mathbf{y}_i(j) \neq \mathbf{h}_i(j)) , \qquad (1)$$

where $I()$ is the indicator function. Hamming Loss should be minimized: if the hamming loss is 0, there is no error in the predicted label vector;

- One error: this is the fraction of instances whose most confident label is incorrect. Because this is an error, the indicator should be minimized:

$$\text{OneError}(F) = \frac{1}{m}\sum_{i=1}^{m} I\left(\mathbf{h}_i\left(\underset{j}{\arg\max}\, f_i\right) \neq \mathbf{y}_i\left(\underset{j}{\arg\max}\, f_i\right)\right) . \qquad (2)$$

- Ranking Loss: this is the average fraction of reversely ordered label pairs for each instance. It can be obtained from the confidence value considering the number of confidence couples correctly ranked (i.e., a true label is ranked before a wrong label). Ranking loss should be minimized.

- Coverage: this is the average number of steps needed to move down the ranked label list of an instance so as to cover all its relevant labels. Coverage should be minimized.

- Average precision: this is the average fraction of relevant labels ranked higher than a particular label. Average precision should be maximized.

Other indicators usually adopted [57] are:

- Aiming: this is the ratio of correctly predicted labels over the practically predicted labels:

$$\text{Aiming}(H) = \frac{1}{m}\sum_{i=1}^{m} \frac{\|\mathbf{h}_i \cap \mathbf{y}_i\|}{\|\mathbf{h}_i\|} . \qquad (3)$$

- Recall: this is the rate of the correctly predicted labels over the actual labels:

$$\text{Recall}(H) = \frac{1}{m}\sum_{i=1}^{m} \frac{\|\mathbf{h}_i \cap \mathbf{y}_i\|}{\|\mathbf{y}_i\|} . \qquad (4)$$

- Accuracy: this is the average ratio of correctly predicted labels over the total labels:

$$\text{Accuracy}(H) = \frac{1}{m}\sum_{i=1}^{m} \frac{\|\mathbf{h}_i \cap \mathbf{y}_i\|}{\|\mathbf{h} \cup \mathbf{y}_i\|} . \qquad (5)$$

- Absolute true: this is the ratio of the perfectly correct prediction events over the total number of prediction events:

$$\text{AbsTrue}(H) = \tfrac{1}{m}\sum_{i=1}^{m} I(\boldsymbol{h}_i = \mathbf{y}_i)\ . \tag{6}$$

- Absolute false: this is the ratio of the completely wrong predictions over the total number of prediction events:

$$\text{AbsFalse}(H) = \tfrac{1}{m}\sum_{i=1}^{m} \frac{\|\boldsymbol{h}_i \cup \mathbf{y}_i\| - \|\boldsymbol{h}_i \cap \mathbf{y}_i\|}{l}\ . \tag{7}$$

Each of the indicators listed above are inside the range [0-1] and should be maximized, except for Absolute false.

# 3  Proposed Approach

In this section, we detail the method for multilabel classification proposed in this work.

## 3.1  Model architecture

We create a Deep Neural Network (DNN) architectures based on GRU and TCN adapted to multilabel classification. The base schema of each base model is provided in Figure 1. The GRU with $N$ hidden units (set to 50 in our experiments) is followed by a max pooling layer and a fully connected layer. The output layer is a sigmoid that provides multiclass classification. The TCN approach has a similar architecture but with max pooling following the fully connected layer. We named these two architectures as GRU_A and TCN_A.

Moreover, experiments showed that both GRU and TCN sometimes perform better if a convolutional level is applied immediately before the network itself. In fact, convolution modifies input features by executing simple mathematical operations with other local features. This may help the model generalize better by consenting features to achieve a higher special independence. We named the TCN based topology with a convolutional level applied immediately before the network itself as TCN_B.

Batch-normalization is a technique for training very deep neural networks that standardizes the inputs to a layer for each mini-batch. This has the effect of stabilizing the learning process and dramatically reducing the number of training epochs required to train deep networks. For this reason, a Batch-Normalization layer is applied to GRU, immediately after convolutional layer discussed above (we named this topology as GRU_B).

Finally, we tested also a sequential combination of GRU_A (without the pooling layer) and TCN_A (i.e. the sigmoid output of GRU_A is the input of TCN_A), named it GRU_TCN.

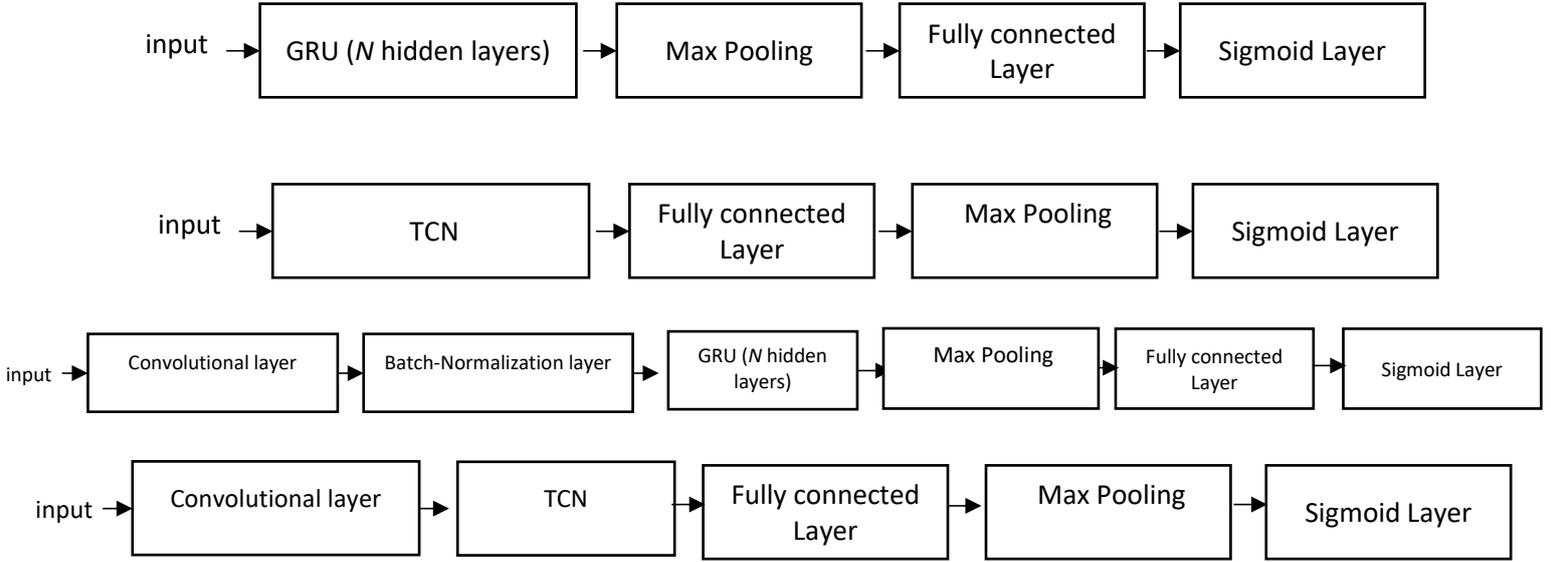

**Figure 1**. Schema of our recurrent DNNs: GRU_A; TCN_A; GRU_B, TCN_B.

The loss function is the binary cross entropy loss between the outputs (predicted labels) and the target or actual labels. The binary cross entropy loss calculates the loss of a set of $m$ observations by computing the following average:

$$\text{CELoss} = -\frac{1}{m}\sum_{i=1}^{m}\sum_{j=1}^{l} \mathbf{y}_i(j) \cdot \log(\mathbf{h}_i(j)) + (1 - \mathbf{y}_i(j)) \cdot \log(1 - \mathbf{h}_i(j)), \qquad (8)$$

where $\mathbf{y}_i \in \{0, 1\}^l$ and $\mathbf{h}_i \in \{0, 1\}^l$ are the actual and predicted label vectors of each sample ($i \in 1...m$), respectively.

## 3.2   Preprocessing

Most of the tested problems/samples require no preprocessing before being classified by the proposed networks. Some preprocessing is needed, however, e.g. when the feature vectors assume values with high variance. The following preprocessing methods are applied:

- Feature normalization in the range [0,1] for the data set ATC_f (this is also a requirement for IMCC [7])
- For the highly sparse data sets (Liu, Arts, bibtex, enron and health), dimensionality reduction is performed by PCA retaining 99% of the variance. This preprocessing is performed only for our networks (not for IMCC), since those datasets are very sparse, proposed networks obtain low performance when using the original data as input.

### 3.3  Gated Recurrent Units (GRU)

As mentioned in the introduction, GRUs [42], like LTSM, are recurrent neural networks with a gating mechanism. GRUs increase the length of term dependencies from the input and handle the gradient vanishing problem. With respect to LTSM, GRU has a forget gate that lets the network decide which part of the old information is relevant to understand the new information [58]. Nonetheless, GRU has fewer parameters than LSTM because there is no output gate. The performance of GRU is similar to LSTM on speech signal modeling, polyphonic music modeling, and natural language processing [9, 59]. According to [60] and [61], however, GRUs tend to perform better on smaller data sets and appear to work well on denoising tasks [62].

The basic components of GRU are a reset gate and an update gate: the first determines how much old information to forget, and the second decides which information should be forgotten and which information should pass on to the output.

Let $x_t$ be the input sequence and initialize $h_0 = 0$. The update gate vector $z_t$ and the reset gate vector $r_t$ can be defined as

$$z_t = \sigma(W_z x_t + U_z h_{t-1} + b_z) \text{ and} \#(9)$$

$$r_t = \sigma(W_r x_t + U_r h_{t-1} + b_r), \quad \#(10)$$

where $W_z, U_z, b_z, W_r, U_r$ and $b_r$ are matrices and vectors and $\sigma$ is the sigmoid function. We define

$$\hat{h}_t = \phi(W_h x_t + U_h(r_t \odot h_{t-1}) + b_h)\#(11)$$

to be the candidate activation vector, where $\phi$ is the tanh activation, and $\odot$ is the Hadamard (component-wise) product. Notice that the term $r_t$ determines the amount of past information that is relevant for the candidate activation vector. The output vector is

$$h_t = (1 - z_t) \odot h_{t-1} + z_t \odot \hat{h}_t. \#(12)$$

The update gate vector $z_t$ measures the amount of old and new information to keep.

## 3.4 Temporal Convolutional Neural Networks (TCN)

TCNs [63] are a class of neural networks that exploit a hierarchy of convolutions to perform fine-grained detection of events in sequence inputs. The most relevant characteristic of TCNs is the one-dimensional convolutions over time that are stacked together to create a deep network. Subsequent layers are padded to have the same size. The convolutions of each layer have a dilation factor that increases exponentially over the layers, letting the first layers look for short-term connections in the data; the deeper layer spots longer-term dependencies based on the features extracted by previous layers. This characteristic allows TCNs to have a large receptive field, overcoming one of the well-known limitations of most RNN architectures.

TCN architectures can vary greatly. The building block used here is composed of a convolution of size three with 175 different filters, followed by a ReLU and batch normalization, followed again by a convolution with the same parameters, a ReLU, and batch normalization. We stacked four blocks to create our model. The dilation factors of the convolutions are $2^{k-1}$ where $k$ is the number of a layer. On top, we use a fully connected layer followed by a max pooling layer; this is followed by the output layer, which is a sigmoid layer for multiclass classification. For training, we use dropout with a probability of 0.05.

### 3.5 IMCC

IMMC [7] is a two-step process: 1) the generation of virtual examples for augmenting the training set and 2) multilabel training. As step 1, augmentation, is what gives IMCC its main performance advantage, it will be the focus of this section. Augmentation is performed via clustering using $k$-means [64] and the calculation of clustering centers. Given the feature matrix $\mathbf{X} = [x_1, x_2...x_n]^T \in \mathbb{R}^{n \times d}$ and the label matrix $\mathbf{Y} = [y_1, y_2...xy_n]^T \in (-1, +1)^{n \times q}$, where $n$ is the number of samples, then, if all samples are partitioned into $c$ clusters $\{\mathcal{Z}_1, \mathcal{Z}_2...\mathcal{Z}_c\}$, and $x_i$ is partitioned into cluster $\mathcal{Z}_j$ so that $x_i \in \mathcal{Z}_j$, it can be assumed that the average of samples will encapsulate the semantic meanings of the cluster. Thus, the center $z_j$ of each cluster $\mathcal{Z}_i$ can be defined as

$$z_j = \frac{1}{|z_j|} \sum_{i=1}^{n} x \cdot \|(x_i \in \mathcal{Z}_j), \tag{13}$$

where $\|(\cdot)$ is an indicator function that is equal to 1 when $x_i \in \mathcal{Z}_j$ or to 0, otherwise.

A complementary training set $\mathcal{D}' = \{z_j, t_j\}_{j=1}^{c}$ can be built by averaging the label vectors of all instances of $\mathcal{Z}_i$ so that

$$t_j = \frac{1}{|z_j|} \sum_{i=1}^{n} y_i \cdot \|(x_i \in \mathcal{Z}_j). \tag{14}$$

The reader is refered to [7] for details on how the objective function handles the original data set $\mathcal{D}$ and the complementary data set $\mathcal{D}'$. In this work, the hyperparameters of IMCC are chosen by five-fold cross validation on the training set.

### 3.6 Pooling

A pooling layer is inserted after the GRU/TCN block to reduce the dimensionality of the processed data by keeping only the most important information and for the purpose of decreasing the probability of overfitting. We have used a single max pooling layer along the time dimension.

## 3.7 Fully connected Layer and Sigmoid Layer

The fully connected layer is comprised of $l$ neurons ($l$ is the number of output labels in a given problem) fully connected with the previous layer. A sigmoid function is used as the activation of this final layer in order to report the activations in the range [0…1], which can be interpreted as the final probabilities (i.e., confidence relevance) of each label. Therefore, the output of the model is a multilabel classification vector, where the output of each neuron of the fully connected layer provides a score (ranging from 0 to 1) for a single label.

## 3.8 Training

Training is performed using the different Adam variants as the optimizer, as detailed in section 4. We use a high learning rate of 0.01 and specify gradient decay and squared gradient decay factors of 0.5 and 0.999, respectively.

Moreover, we clip the gradients with a threshold of one using L2 norm gradient clipping. The minibatch size has been fixed to 30, while the number of epochs in our experiments is set to 150 for GRU and 100 for TNC.

## 3.9 Ensemble creation

Ensembles combine the output of multiple models to improve system performance and to prevent overfitting. As noted in the introduction, prediction and generalization of an ensemble is enhanced when the diversity of the classifiers is increased. Our ensemble architecture is based on the fusion via average rule of several models trained on the same problem. Random initialization is a simple way for generating diversity in the neural network models but is significantly enhanced with the application of different optimization strategies. Optimizers play a fundamental role in finding a minimum of the fitness function; different optimization strategies can converge to different local minima and thus achieve different optima. We evaluate several optimizers (presented in section 4), which are suitable for ensemble creation: Adam optimizer [44], diffGrad [65] and four variants of this approach: DGrad [66], Cos#1 [66], Exp [67], and Sto (new).

We build an ensemble of 40 networks as follows: for each layer of each network, we randomly choose which optimization approach (DGrad, Cos#1, Exp, or Sto) to use in that layer; in this way, a set of 40 models are built.

## 4 Optimizers

In this section, the optimizer methods underlying our approach for ensemble creation are explained.

### 4.1 Adam optimizer

Adam is an optimizer introduced in [44] that computes adaptive learning rates for each parameter combining the ideas of momentum and adaptive gradient. The update rule for Adam is based on the value of the gradient at the current step and on the exponential moving averages of the gradient and its square. To be more precise, Adam defines the moving averages $m_t$ (the first moment) and $u_t$ (the second moment) as

$$m_t = \rho_1 m_{t-1} + (1 - \rho_1) g_t \quad \#(15)$$

$$u_t = \rho_2 u_{t-1} + (1 - \rho_2) g_t^2 \quad \#(16)$$

where $g_t$ is the gradient at time $t$, the square on $g_t$ stands for the component-wise square, $\rho_1$ and $\rho_2$ are hyperparameters representing the exponential decay rate for the first moment and the second moment estimates (usually set to 0.9 and 0.999, respectively), with moments initialized to 0: $m_o = u_0 = 0$. In order to take into account the fact that the value of moving averages will be very small due to their initialization to zero (especially in the first steps), the authors of Adam define a bias-corrected version of the moving averages:

$$\widehat{m}_t = \frac{m_t}{(1 - \rho_1^t)} \quad \#(17)$$

$$\widehat{u}_t = \frac{u_t}{(1-\rho_2^t)} . \#(18)$$

The final update for each $\theta_t$ parameter of the network is

$$\theta_t = \theta_{t-1} - \lambda \frac{\widehat{m}_t}{\sqrt{\widehat{u}_t} + \epsilon} , \#(19)$$

where $\lambda$ is the learning rate, $\epsilon$ is a small positive number to prevent division by zero (usually set to $10^{-8}$), and all operations are component-wise.

Notice that while $g_t$ might have positive or negative components, $g_t^2$ has only positive components. Hence, if the gradient changes sign often, the value of $\widehat{m}_t$ might be much lower than $\sqrt{\widehat{u}_t}$. In this case, the step size is very small.

## 4.2 diffGrad optimizer

The diffGrad optimizer is a method proposed in [65] that takes into account the difference of the gradient in order to set the learning rate. Based on the observation that gradient changes begin to reduce during training and that this is often indicative of the presence of a global minima, diffGrad applies an adaptive adjustment driven by the difference between the present and the immediate past values to lock parameters into a global minimum. Therefore, the step size is larger for faster gradient changing and smaller for lower gradient changing parameters. In order to define the update function, the absolute difference of two consecutive steps of the gradient is defined as:

$$\Delta g_t = |g_{t-1} - g_t| \#(20)$$

The final update for each $\theta_t$ parameter of the network is as in equation (9) where $\widehat{m}_t$ and $\widehat{u}_t$ are defined as in equations (15) and (16), and the learning rate is modulated by the Sigmoid of $\Delta g_t$:

$$\xi_t = Sig(\Delta g_t) \#(21)$$

$$\theta_{t+1} = \theta_t - \lambda \cdot \xi_t \frac{\widehat{m}_t}{\sqrt{\widehat{u}_t} + \epsilon} . \#(22)$$

## 4.3 diffGrad Variants

In this work, we evaluate the following variants of the diffGrad optimization method:

- DGrad [66] is based on the moving average of the element-wise squares of the parameter gradients;
- Cos#1 [66] is a minor variant of DGrad based on the application of a cyclic learning rate [68];
- Exp (NEW), it is a variant of a method proposed in [67], it is based on the application of an exponential function;
- Sto (NEW) is a stochastic approach to set the learning rate with the aim of stalling the optimizer on flat surfaces or small wells.

The proposed approaches have different methods for defining $\xi_t$, followed by the application of equation (16) to calculate the final update for $\theta_t$.

**DGrad** [66] takes up the ideas of diffGrad by defining the following absolute difference between two consecutive steps of the gradient as

$$\Delta ag_t = |g_t - avg_t| \,, \#(23)$$

where $avg_t$ contains the moving average of the element-wise squares of the parameter gradients; we then normalize $\Delta ag_t$ by its maximum as

$$\widehat{\Delta ag}_t = \left(\frac{\Delta ag_t}{\max(\Delta ag_t)}\right). \#(24)$$

We define $\xi_t$ as the Sigmoid of $4 \cdot \widehat{\Delta ag}_t$, where the rationale of the multiplication by "4" is to increase the range of the output of the sigmoid function:

$$\xi_t = Sig(4 \cdot \widehat{\Delta ag}_t) \,. \#(25)$$

**Cos#1** [66] is a variant of DGrad which exploits the idea of using a cyclic learning rate, with the aim of improving classification accuracy without tuning and with fewer iterations [18].

We use the *cos*() periodic function to define a range of variation of the learning rate according to the following formulation:

$$lr_t = \left(2 - \left|\cos\left(\frac{\pi \cdot t}{steps}\right)\right| e^{-0.01 \cdot (mod(t, steps)+1)}\right), \#(26)$$

where *mod*() denotes the function modulo and where *steps*=30 is the period. The plot of $lr_t$ for *t* in the range 1:2×*steps* is reported in Figure 2.

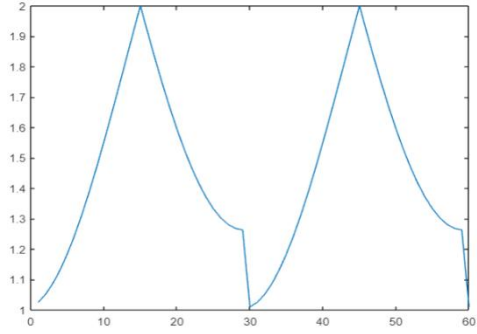

**Figure 2**. Cyclic learning rate.

In this variant, $lr_t$ is used as a multiplier of $\Delta \widehat{ag}_t$ in the definition of $\xi_t$ (equation (21)), which becomes:

$$\xi_t = Sig(4 \cdot lr_t \cdot \Delta \widehat{ag}_t) . \#(27)$$

**Exp** [67] consists of two simple element-wise operations: product and exponential. The purpose of this formulation is to mitigate the effect of large variations in the gradient, but it also allows the function to converge for small values. The function in eq. (28) has a pattern that decays slower than the negative exponential for larger values; moreover, thanks to normalization, it gives less focus on gradient variations that tend to zero, widening the area of greater gain, thus:

$$lr_t = \Delta ag_t \cdot e^{(-k \cdot \Delta ag_t)}, \#(28)$$

with *k* fixed at 2 in this work.

The final learning rate ($\xi_t$) is the normalized result multiplied by a correction factor (1.5), which helps to move the mean towards the unit (see the plot reported in Figure 3):

$$\xi_t = 1.5 \frac{lr_t}{\max(lr_t)} . \#(29)$$

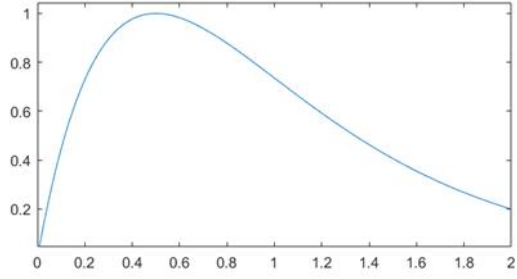

**Figure 3.** Plot of equation (26).

**Sto** (the abbreviation of Stochastic) is our new variant that is designed to reduce the possibility that the optimizer will stall on flat surfaces or small wells by adding noise to the learning rate. Additive noise is independent of the gradient direction; therefore, it adds a degree of uncertainty to the search for the optimum that could help find local minimum when the optimizer is struggling due to too large a learning rate.

Let $\mathcal{X}$ be a matrix of independent uniform random variables in range [0,1] and $J$ an all-ones matrix:

$$\mathcal{X} = \begin{bmatrix} X_{1,1} & \cdots & X_{1,n} \\ \vdots & \ddots & \vdots \\ X_{m,1} & \cdots & X_{m,n} \end{bmatrix} \quad \text{and} \quad J = \begin{bmatrix} 1 & \cdots & 1 \\ \vdots & \ddots & \vdots \\ 1 & \cdots & 1 \end{bmatrix},$$

where $X_{i,j} \sim \mathcal{U}(0,1)$ are random variables with a uniform density function. The learning rate is defined as

$$lr_t = \Delta ag_t \cdot e^{(-4 \cdot \Delta ag_t)} \cdot (\mathcal{X} + 0.5 \cdot J) \#(30)$$

$$\xi_t = 1.5 \frac{lr_t}{\max(lr_t)}. \#(31)$$

In eq. 30, the matrix $J$ is used to shift $X$'s range by 0.5 in order to move the mean across 1.

## 5 Experimental Results

In the first experiment (see Table 3), the aim was to build and evaluate different variants of the base models combined with the components described in previous sections, with all classifiers fused by the average rule. A descriptive summary of each of these ensembles in terms of the number of classifiers and hidden units in the base GRU and TCN models is provided in Table 2, along with the number of training epochs and the loss function. Starting from a standalone GRU_A with 50 hidden units trained by Adam for 50 epochs (labeled Adam_sa), we generated new ensembles by incrementally increasing the complexity of the base GRU by fusing 10 Adam_sa (Adam_10s), increasing the number of epochs (Adam_10), changing the optimizer (DG_10, Cos_10, Exp_10, Sto_10), and fusing together the best results as follows:

- DG_Cos is the fusion of DG_10 + Cos_10;
- DG_Cos_Exp is the fusion of DG_10 + Cos_10 + Exp_10;
- DG_Cos_Exp_Sto is the fusion of DG_10 + Cos_10 + Exp_10 + Sto_10;
- StoGRU is an ensemble composed of 40 GRU_A, combined by average rule, as explained in section 3.10;
- StoGRU_B as StoGRU but based on GRU_B;
- StoTCN is an ensemble of 40 TCN_A, combined by average rule, as explained in section 3.10;
- StoTCN_B as StoTCN but based on TCN_B;
- StoGRU_TCN is an ensemble of 40 GRU_TCN each coupled with the stocastic approach explained in section 3.10;
- ENNbase is the fusion by average rule of StoGRU and StoTCN.
- ENN is the fusion by average rule of StoGRU, StoTCN, StoGRU_B, StoTCN_B and StoGRU_TCN.

In Table 4, we report the performance of the ensembles listed in Table 2 and compare them with IMCC [7]. For IMCC the parameters were chosen by five-fold cross-validation on the training set. We also report two different fusions between ENN and IMCC:

- ENN+IMCC is the sum rule between ENN and IMCC; before fusion the scores of ENN were normalized since it has a different range of values compared to IMCC. Normalization was performed as ENN=(ENN-0.5)×2 with the classification threshold set to zero;
- ENN+3×IMCC is the same as the previous fusion, but the scores of IMCC are weighted by a factor of three.

| Name | Hidden units | #Classifiers | #Epoch | Optimizer |
|---|---|---|---|---|
| Adam_sa | 50 | 1 | 50 | Adam |
| Adam_10s | 50 | 10 | 50 | Adam |
| Adam_10 | 50 | 10 | 150 | Adam |
| DG_10 | 50 | 10 | 150 | DGrad |
| Cos_10 | 50 | 10 | 150 | Cos |
| Exp_10 | 50 | 10 | 150 | Exp |
| Sto_10 | 50 | 10 | 150 | Sto |
| DG_Cos= DG_10 + Cos_10 | 50 | 20 | 150 | DGrad,Cos |
| DG_Cos_Exp = DG_10 + Cos_10 + Exp_10 | 50 | 30 | 150 | DGrad,Cos, Exp |
| DG_Cos_Exp_Sto = DG_10 + Cos_10 + Exp_10 + Sto_10 | 50 | 40 | 150 | DGrad,Cos, Exp, Sto |
| StoGRU | 50 | 40 | 150 | DGrad,Cos, Exp, Sto |
| StoGRU_B | 50 | 40 | 150 | DGrad,Cos, Exp, Sto |
| StoTCN | --- | 40 | 100 | DGrad,Cos, Exp, Sto |
| StoTCN_B | --- | 40 | 100 | DGrad,Cos, Exp, Sto |
| StoGRU_TCN | 50 (GRU) | 40 | 150 | DGrad,Cos, Exp, Sto |
| ENNbase=StoTCN+StoGRU | 50 (GRU) | 80 | 100 (TCN) / 150 (GRU) | DGrad,Cos, Exp, Sto |
| ENN=StoTCN+StoGRU+StoTCN_B+StoGRU_B+StoGRU_TCN | 50 (GRU, GRU_B & GRU_TCN) | 200 | 100 (TCN & TCN_B) / 150 (GRU, GRU_B & GRU_TCN) | DGrad,Cos, Exp, Sto |

**Table 2.** Descriptive summary of tested ensembles.

In Table 3 we report an ablation study for assessing the performance improvement that each module of our ensemble allows to obtain; the topology named GRU_A has been used in these tests, anyway, similar conclusions can be obtained using the others four topologies tested in this work. In table 3, we compare approaches using Wilcoxon signed rank test.

| Method | Comparison |
|---|---|
| Adam_10s | Outperforms Adam_sa with a p-value of 0.0156 |
| Adam_10 | Outperforms Adam_sa with a p-value of 0.0172<br>Same performance of Adam_10s |
| DG_10 | Outperforms Adam_10 with a p-value of 0.0064 |
| Cos_10 | Outperforms Adam_10 with a p-value of 0.0137 |
| Exp_10 | Outperforms Adam_10 with a p-value of 0.0016 |
| Sto_10 | Outperforms Adam_10 with a p-value of 0.1014 |
| DG_Cos_Exp_Sto | Outperforms Exp_10 (the best of the approaches reported above in this table) with a p-value of 0.0134 |
| StoGRU | Outperforms DG_Cos_Exp_Sto with a p-value of 0.0922 |

**Table 3.** Ablation study show that StoGRU is the best among the GRU_A based approaches.

In Table 4, the performance of each method in the twelve data sets is reported in terms of average precision.

| Average Precision | Cal500 | image | scene | yeast | arts | ATC | ATC_f | Liu | mAn | bibtex | enron | health | Average |
|---|---|---|---|---|---|---|---|---|---|---|---|---|---|
| IMCC | **0.502** | 0.836 | 0.904 | **0.773** | 0.619 | 0.866 | **0.922** | 0.523 | 0.978 | **0.623** | 0.714 | 0.781 | 0.753 |
| StoGRU | 0.498 | 0.851 | 0.911 | 0.740 | 0.561 | 0.872 | 0.872 | 0.485 | **0.979** | 0.403 | 0.680 | 0.739 | 0.715 |
| StoGRU_B | 0.490 | **0.861** | 0.908 | 0.741 | 0.555 | 0.877 | 0.848 | 0.485 | 0.978 | 0.400 | 0.688 | 0.724 | 0.712 |
| StoTCN | 0.498 | 0.847 | 0.913 | 0.764 | 0.506 | 0.882 | 0.900 | 0.498 | 0.977 | 0.406 | 0.669 | 0.710 | 0.714 |
| StoTCN_B | 0.497 | 0.855 | 0.917 | 0.765 | 0.541 | 0.883 | 0.903 | 0.505 | 0.976 | 0.404 | 0.666 | 0.732 | 0.720 |
| StoGRU_TCN | 0.491 | 0.852 | 0.916 | 0.752 | 0.592 | 0.890 | 0.913 | 0.510 | 0.977 | 0.354 | 0.674 | 0.764 | 0.724 |
| ENNbase | **0.502** | 0.855 | 0.922 | 0.756 | 0.552 | 0.888 | 0.916 | 0.497 | **0.979** | 0.417 | 0.687 | 0.735 | 0.726 |
| ENN | 0.499 | 0.859 | **0.924** | 0.762 | 0.582 | **0.893** | 0.916 | 0.505 | 0.979 | 0.424 | 0.689 | 0.749 | 0.732 |
| ENN+IMCC | 0.502 | **0.856** | **0.922** | **0.783** | **0.631** | 0.883 | **0.927** | 0.521 | 0.979 | 0.622 | 0.714 | 0.783 | **0.760** |
| ENN+3×IMCC | **0.503** | 0.845 | 0.917 | 0.777 | 0.627 | 0.876 | 0.926 | **0.524** | **0.979** | **0.624** | **0.716** | **0.784** | 0.758 |

**Table 4.** Average precision of the tested ensembles and state of the art in twelve data sets (boldface values indicate the best performance within each group of similar approaches).

The following conclusions can be drawn from Tables 3/4:

- The ensemble of GRU greatly outperforms stand-alone GRU (Adam_sa vs Adam_10s);

- The new variants of Adam outperform the original Adam, and their ensemble performs better than IMCC on some data sets;
- GRU/TCN-based methods work poorly in largely sparse data sets (i.e., Arts, Liu, bibtex, enron and health);
- StoGRU_TCN outperforms the other ensembles based on GRU/TCN; StoGRU and StoTCN perform similarly;
- Among the four Adam variants, there is no clear winner;
- ENN outperforms each method that built it;
- The fusions between ENN and IMCC produce the best results: ENN+3×IMCC tops IMCC in all the data sets (instead ENN+IMCC has performance equal or lower to IMCC in some datasets), so it is our suggested approach.

In the tests that follow, see Tables 5-7, we simplify the label of our best ensemble *ENN+3×IMCC* to *Ens* to reduce clutter. In Table 5, we compare IMCC and Ens using additional performance indicators. It can be observed in examining this table that the proposed ensemble outperforms IMCC.

|  | One Error ↓ | Hamming Loss ↓ | Ranking Loss ↓ | Coverage ↓ | Avg Precision ↑ |
|---|---|---|---|---|---|
| Cal500-IMCC | **0.150** | 0.134 | 0.182 | 0.736 | 0.502 |
| Cal500-Ens | **0.150** | 0.134 | **0.180** | **0.732** | **0.503** |
| Image-IMCC | 0.237 | 0.150 | 0.138 | 0.173 | 0.836 |
| Image-Ens | **0.230** | **0.147** | **0.127** | **0.164** | **0.845** |
| scene-IMCC | 0.164 | 0.070 | 0.053 | 0.062 | 0.904 |
| scene-Ens | **0.141** | **0.064** | **0.046** | **0.056** | **0.917** |
| Yest-IMCC | 0.220 | 0.185 | 0.165 | 0.448 | 0.773 |
| Yest-Ens | **0.215** | **0.178** | **0.158** | **0.437** | **0.777** |
| Arts-IMCC | 0.438 | **0.054** | 0.164 | 0.242 | 0.619 |
| Arts-Ens | **0.435** | 0.054 | **0.149** | **0.225** | **0.627** |
| Bibtex-IMCC | 0.336 | 0.012 | 0.079 | 0.158 | 0.623 |
| Bibtex-Ens | **0.336** | **0.012** | **0.074** | **0.147** | **0.624** |
| Enron-IMCC | 0.226 | **0.044** | 0.072 | 0.211 | 0.714 |
| Enron-Ens | **0.223** | 0.044 | **0.070** | **0.207** | **0.716** |
| Health-IMCC | **0.266** | 0.035 | 0.052 | 0.107 | 0.781 |
| Health-Ens | **0.266** | **0.035** | **0.048** | **0.100** | **0.784** |

**Table 5**. Comparison using more performance indicators.

In table 6, we compare our best approach with the state of the art on the mAn data set using the performance measures aiming, coverage, accuracy, absolute true, and absolute false.

| mAn | Aiming | Coverage | Accuracy | Absolute True | Absolute False |
|---|---|---|---|---|---|
| [4] | 88.31 | 85.06 | 84.34 | 78.78 | 0.07 |
| [6] | 96.21 | 97.77 | 95.46 | 92.26 | 0.00 |
| IMCC | 92.80 | 92.02 | 88.83 | 80.76 | 1.43 |
| Ens | 93.59 | 92.94 | 89.94 | 83.03 | 1.25 |

**Table 6.** Performance in the mAn data set.

In the mAn data set, Ens outperforms IMCC using these measures, but Ens does not achieve state-of-the-art performance; the recent ad hoc method proposed in [6] performs better on this data set than Ens does.

Finally, in Table 7, the proposed ensemble is compared with the literature across the twelve data sets using average precision as the performance indicator. As can be observed, Ens obtains state-of-the-art performance using this measure on this collection of data sets.

| Average Precision | Cal500 | image | scene | yeast | arts | ATC | ATC_f | Liu | mAn | bibtex | enron | health |
|---|---|---|---|---|---|---|---|---|---|---|---|---|
| Ens | 0.503 | **0.845** | **0.917** | 0.777 | **0.627** | **0.876** | **0.926** | **0.524** | **0.979** | 0.624 | 0.716 | **0.784** |
| FastAi [40] | 0.425 | 0.824 | 0.899 | 0.718 | 0.588 | 0.860 | 0.908 | 0.414 | 0.976 | --- | --- | --- |
| IMCC | 0.502 | 0.836 | 0.904 | 0.773 | 0.619 | 0.866 | 0.922 | 0.523 | 0.978 | 0.623 | 0.714 | 0.781 |
| hML[20] | 0.453 | 0.810 | 0.885 | **0.792** | 0.538 | 0.831 | 0.854 | 0.433 | 0.965 | --- | --- | --- |
| ECC [7] | 0.491 | 0.797 | 0.857 | 0.756 | 0.617 | --- | --- | --- | --- | 0.617 | 0.657 | 0.719 |
| MAHR [7] | 0.441 | 0.801 | 0.861 | 0.745 | 0.524 | --- | --- | --- | --- | 0.524 | 0.641 | 0.715 |
| LLSF [7] | 0.501 | 0.789 | 0.847 | 0.617 | **0.627** | --- | --- | --- | --- | **0.627** | 0.703 | 0.780 |
| JFSC [7] | 0.501 | 0.789 | 0.836 | 0.762 | 0.597 | --- | --- | --- | --- | 0.597 | 0.643 | 0.751 |
| LIFT [7] | 0.496 | 0.789 | 0.859 | 0.766 | **0.627** | --- | --- | --- | --- | **0.627** | 0.684 | 0.708 |
| [17] | --- | --- | --- | --- | --- | --- | --- | 0.513 | --- | --- | --- | --- |
| [69] | --- | --- | --- | --- | --- | --- | --- | 0.261 | --- | --- | --- | --- |
| hMuLab [20] | --- | --- | --- | 0.778 | --- | --- | --- | --- | --- | --- | --- | --- |
| MlKnn [20] | --- | --- | --- | 0.766 | --- | --- | --- | --- | --- | --- | --- | --- |
| RaKel [20] | --- | --- | --- | 0.715 | --- | --- | --- | --- | --- | --- | --- | --- |
| ClassifierChain [20] | --- | --- | --- | 0.624 | --- | --- | --- | --- | --- | --- | --- | --- |
| IBLR [20] | --- | --- | --- | 0.768 | --- | --- | --- | --- | --- | --- | --- | --- |
| MLDF [70] | **0.512** | 0.842 | 0.891 | 0.770 | --- | --- | --- | --- | --- | --- | **0.742** | --- |
| RF_PCT [70] | **0.512** | 0.829 | 0.873 | 0.758 | --- | --- | --- | --- | --- | --- | 0.729 | --- |
| DBPNN [70] | 0.495 | 0.672 | 0.563 | 0.738 | --- | --- | --- | --- | --- | --- | 0.679 | --- |
| MLFE [70] | 0.488 | 0.817 | 0.882 | 0.759 | --- | --- | --- | --- | --- | --- | 0.656 | --- |
| ECC [70] | 0.482 | 0.739 | 0.853 | 0.752 | --- | --- | --- | --- | --- | --- | 0.646 | --- |
| RAKEL [70] | 0.353 | 0.788 | 0.843 | 0.720 | --- | --- | --- | --- | --- | --- | 0.596 | --- |
| [16] | --- | --- | --- | 0.758 | --- | --- | --- | --- | --- | --- | --- | --- |
| [71] | 0.484 | --- | --- | 0.740 | --- | --- | --- | --- | --- | --- | --- | --- |

**Table 7.** Other comparisons with the literature using average precision.

## Conclusion

In this paper, we have proposed a novel system for multilabel classification that combines ensembles of gated recurrent units (GRU) with temporal convolutional neural networks (TCN) trained with variants of Adam optimization (with one new variant proposed here for the first time). This approach is also combined with Incorporating Multiple Clustering Centers (IMCC) for superior multilabel classification. To validate our approach, it is tested across a set of twelve benchmark data sets representing many different applications. Experimental results show that the proposed ensemble obtains state-of-the-art performance.

Future studies will focus on the fusion of this approach with other topologies for extracting features.

## Acknowledgement

Through their GPU Grant Program, NVIDIA donated the TitanX GPU that was used to train the CNNs presented in this work.

## References

[1] E. G. Galindo and S. Ventura, "Multi label learning: a review of the state of the art and ongoing research," *Wiley Interdisciplinary Reviews: Data Mining and Knowledge Discovery,* vol. 4, no. 6, pp. 411-444, 2014, doi: doi.org/10.1002/widm.1139.

[2] A. Elisseeff and J. Weston, *A kernel method for multi-labelled classification* (NIPS). MIT Press Direct, 2001.

[3] L. Nanni, A. Lumini, and S. Brahnam, "Neural networks for Anatomical Therapeutic Chemical (ATC)," *ArXiv,* vol. abs/2101.11713, 2021.

[4] X. Cheng, W.-Z. Lin, X. Xiao, and K.-C. Chou, "pLoc_bal-mAnimal: predict subcellular localization of animal proteins by balancing training dataset and PseAAC," *Bioinformatics,* vol. 35, no. 3, pp. 398-406, 2018, doi: 10.1093/bioinformatics/bty628.

[5] L. Chen *et al.*, "Predicting gene phenotype by multi-label multi-class model based on essential functional features," *Molecular genetics and genomics : MGG,* vol. 296, no. 4, pp. 905-918, 2021, doi: 10.1007/s00438-021-01789-8.

[6] Y. Shao and K. Chou, "pLoc_Deep-mAnimal: A Novel Deep CNN-BLSTM Network to Predict Subcellular Localization of Animal Proteins," *Natural Science,* vol. 12, 5, pp. 281-291, 2020, doi: 10.4236/ns.2020.125024.

[7] S. Shu *et al.*, "Incorporating Multiple Cluster Centers for Multi-Label Learning," *ArXiv,* vol. abs/2004.08113, 2020.

[8] M. Ibrahim, M. U. G. Khan, F. Mehmood, M. Asim, and W. Mahmood, "GHS-NET a generic hybridized shallow neural network for multi-label biomedical text classification," *Journal of biomedical informatics,* vol. 116, p. 103699, 2021, doi: 10.1016/j.jbi.2021.103699.

[9] M. Ravanelli, P. Brakel, M. Omologo, and Y. Bengio, "Light Gated Recurrent Units for Speech Recognition," *IEEE Transactions on Emerging Topics in Computational Intelligence,* vol. 2, no. 2, pp. 92-102, 2018, doi: 10.1109/TETCI.2017.2762739.

[10] Y. Kim and J. Kim, "Human-Like Emotion Recognition: Multi-Label Learning from Noisy Labeled Audio-Visual Expressive Speech," presented at the 2018 IEEE International Conference on Acoustics, Speech and Signal Processing (ICASSP), 2018.

[11] M. B. Messaoud, I. Jenhani, N. B. Jemaa, and M. W. Mkaouer, "A Multi-label Active Learning Approach for Mobile App User Review Classification," in *KSEM*, 2019.

[12] J. P. Singh and K. Nongmeikapam, "Negative Comments Multi-Label Classification," *2020 International Conference on Computational Performance Evaluation (ComPE),* pp. 379-385, 2020, doi: 10.1109/ComPE49325.2020.9200131.

[13] M. Boutell, J. Luo, X. Shen, and C. M. Brown, "Learning multi-label scene classification," *Pattern Recognit.,* vol. 37, no. 9, pp. 1757-1771, 2004.

[14] G. Tsoumakas, I. Katakis, and I. Vlahavas, "Mining Multi-label Data," in *Data Mining and Knowledge Discovery Handbook*, 2010: Springer, pp. 667–685.

[15] J. Read, B. Pfahringer, G. Holmes, and E. Frank, "Classifier chains for multi-label classification," *Mach Learn,* vol. 85, pp. 333-359, 2011.

[16] W. Qian, C. Xiong, and Y. Wang, "A ranking-based feature selection for multi-label classification with fuzzy relative discernibility," *Applied Soft Computing,* vol. 102, p. 106995, 2021/04/01/ 2021, doi: https://doi.org/10.1016/j.asoc.2020.106995.

[17] W. Zhang, F. Liu, L. Luo, and J. Zhang, "Predicting drug side effects by multi-label learning and ensemble learning," *BMC Bioinformatics,* vol. 16, no. 365, 2015.

[18] J. Huang, G. Li, Q. Huang, and X. Wu, "Joint Feature Selection and Classification for Multilabel Learning," *IEEE Transactions on Cybernetics,* vol. 48, no. 3, pp. 876-889, 2018.

[19] J. Huang, G. Li, Q. Huang, and X. Wu, "Learning Label-Specific Features and Class-Dependent Labels for Multi-Label Classification," *IEEE Transactions on Knowledge and Data Engineering,* vol. 28, no. 12, pp. 3309-3323, 2016.

[20] P. Wang, R. Ge, X. Xiao, M. Zhou, and F. Zhou, "hMuLab: A Biomedical Hybrid MUlti-LABel Classifier Based on Multiple Linear Regression," *IEEE/ACM Trans. Comput. Biol. Bioinformatics,* vol. 14, no. 5, pp. 1173–1180, 2017, doi: 10.1109/tcbb.2016.2603507.

[21] G. Tsoumakas, I. Katakis, and I. Vlahavas, "Random k-Labelsets for Multilabel Classification," *IEEE Transactions on Knowledge and Data Engineering,* vol. 23, no. 7, pp. 1079-1089, 2011.

[22] Y. Yang and J. Jiang, "Adaptive Bi-Weighting Toward Automatic Initialization and Model Selection for HMM-Based Hybrid Meta-Clustering Ensembles," *IEEE Transactions on Cybernetics,* vol. 49, no. 5, pp. 1657-1668, 2019.

[23] J. M. Moyano, E. G. Galindo, K. Cios, and S. Ventura, "Review of ensembles of multi-label classifiers: Models, experimental study and prospects," *Inf. Fusion,* vol. 44, no. November, pp. 33-45, 2018.

[24] Y. Xia, K. Chen, and Y. Yang, "Multi-label classification with weighted classifier selection and stacked ensemble," *Inf. Sci.,* vol. 557, no. May, pp. 421-442, 2021.

[25] J. M. Moyano, E. G. Galindo, K. Cios, and S. Ventura, "An evolutionary approach to build ensembles of multi-label classifiers," *Inf. Fusion,* vol. 50, no. October, pp. 168-180, 2019.

[26] R. Wang, S. Kwong, X. Wang, and Y. Jia, "Active k-labelsets ensemble for multi-label classification," *Pattern Recognit.,* vol. 109, no. January, p. 107583, 2021.

[27] Imagga. "Imagga website." https://imagga.com/solutions/auto-tagging (accessed 2021).

[28] Clarifai. "Clarifai website." https://www.clarifai.com/ (accessed 2021).
[29] Microsoft. "Computer-vision API website." https://www.microsoft.com/cognitive-services/en-us/computer-vision-api (accessed 2021).
[30] IBM. "Visual Recognition." https://www.ibm.com/watson/services/visual-recognition/ (accessed 2020).
[31] Google. "Google Cloud Vision." https://cloud.google.com/vision/ (accessed 2021).
[32] Wolfram. "Wolfram Alpha: Image Identification Project." https://www.imageidentify.com/ (accessed 2020).
[33] DeepDetect. "DeepDetect." https://www.deepdetect.com/ (accessed 2021).
[34] K. Simonyan and A. Zisserman, "Very deep convolutional networks for large-scale image recognition," Cornell University, arXiv:1409.1556v6 2014.
[35] C. Szegedy, V. Vanhoucke, S. Ioffe, J. Shlens, and Z. Wojna, "Rethinking the inception architecture for computer vision," presented at the IEEE Conference on Computer Vision and Pattern Recognition, 2016.
[36] C. Szegedy, S. Ioffe, V. Vanhoucke, and A. Alemi, "Inception-v4, inception-resnet and the impact of residual connections on learning," in "arxiv.org," Cornell University, https://arxiv.org/pdf/1602.07261.pdf, 2016.
[37] K. He, X. Zhang, S. Ren, and J. Sun, "Deep Residual Learning for Image Recognition," *2016 IEEE Conference on Computer Vision and Pattern Recognition (CVPR),* pp. 770-778, 2015.
[38] M. Sandler, A. Howard, M. Zhu, A. Zhmoginov, and L. Chen, "MobileNetV2: Inverted Residuals and Linear Bottlenecks," in *2018 IEEE/CVF Conference on Computer Vision and Pattern Recognition*, 18-23 June 2018 2018, pp. 4510-4520, doi: 10.1109/CVPR.2018.00474.
[39] J. Redmon and A. Farhadi, "YOLOv3: An Incremental Improvement," *ArXiv,* vol. abs/1804.02767, 2018.
[40] J. Howard and S. Gugger, "Fastai: A Layered API for Deep Learning," *Information,* vol. 11, no. 2, p. 108, 2020. [Online]. Available: https://www.mdpi.com/2078-2489/11/2/108.
[41] A. Kubany, S. B. Ishay, R.-s. Ohayon, A. Shmilovici, L. Rokach, and T. Doitshman, "Comparison of state-of-the-art deep learning APIs for image multi-label classification using semantic metrics," *Expert Syst. Appl.,* vol. 161, no. 15, p. 113656, 2020.
[42] K. Cho *et al.*, "Learning Phrase Representations using RNN Encoder Decoder for Statistical Machine Translation," in *EMNLP*, Varna, Bulgaria, 2014, pp. 25-32.
[43] C. S. Lea, M. D. Flynn, R. Vidal, A. Reiter, and G. Hager, "Temporal Convolutional Networks for Action Segmentation and Detection," presented at the 2017 IEEE Conference on Computer Vision and Pattern Recognition (CVPR), Honolulu, Hawaii, 2017.
[44] D. P. Kingma and J. Ba, "Adam: A Method for Stochastic Optimization," *CoRR,* vol. abs/1412.6980, 2015.
[45] M.-L. Zhang and Z.-H. Zhou, "Multilabel Neural Networks with Applications to Functional Genomics and Text Categorization," *IEEE Transactions on Knowledge and Data Engineering,* vol. 18, no. 10, pp. 1338-1351, 2006, doi: 10.1109/TKDE.2006.162.
[46] R. Stivaktakis, G. Tsagkatakis, and P. Tsakalides, "Deep Learning for Multilabel Land Cover Scene Categorization Using Data Augmentation," *IEEE Geoscience and Remote Sensing Letters,* vol. 16, 7, pp. 1031-1035, 2019.
[47] H. Zhu *et al.*, "Automatic multilabel electrocardiogram diagnosis of heart rhythm or conduction abnormalities with deep learning: a cohort study," *The Lancet. Digital health,* vol. 2, no. 9, pp. e348-e357, 2020.
[48] N. Navamajiti, T. Saethang, and D. Wichadakul, "McBel-Plnc: A Deep Learning Model for Multiclass Multilabel Classification of Protein-lncRNA Interactions," presented at the Proceedings of the 2019 6th International Conference on Biomedical and Bioinformatics Engineering (ICBBE'19), Shanghai, China, 2019.
[49] B. Namazi, G. Sankaranarayanan, and V. Devarajan, "LapTool-Net: A Contextual Detector of Surgical Tools in Laparoscopic Videos Based on Recurrent Convolutional Neural Networks," *ArXiv,* vol. abs/1905.08983, 2019.

[50] X. Zhou, Y. Li, and W. Liang, "CNN-RNN Based Intelligent Recommendation for Online Medical Pre-Diagnosis Support," *IEEE/ACM Transactions on Computational Biology and Bioinformatics,* vol. 18, no. 3, pp. 912-921, 2021.

[51] A. E. Samy, S. R. El-Beltagy, and E. Hassanien, "A Context Integrated Model for Multi-label Emotion Detection," *Procedia Computer Science,* vol. 142, pp. 61-71, 2018/01/01/ 2018, doi: https://doi.org/10.1016/j.procs.2018.10.461.

[52] J. Zhang, Q. Chen, and B. Liu, "NCBRPred: predicting nucleic acid binding residues in proteins based on multilabel learning," *Briefings in bioinformatics,* vol. 22, no. 2, 2021.

[53] D. Li, H. Wu, J. Zhao, Y. Tao, and J. Fu, "Automatic Classification System of Arrhythmias Using 12-Lead ECGs with a Deep Neural Network Based on an Attention Mechanism," *Symmetry,* vol. 12, no. 11, p. 1827, 2020. [Online]. Available: https://www.mdpi.com/2073-8994/12/11/1827.

[54] D. Turnbull, L. Barrington, D. A. Torres, and G. Lanckriet, "Semantic Annotation and Retrieval of Music and Sound Effects," *IEEE Transactions on Audio, Speech, and Language Processing,* vol. 16, no. 22, pp. 467-476, 2008, doi: 10.1109/TASL.2007.913750.

[55] M.-L. Zhang and Z. Zhou, "ML-KNN: A lazy learning approach to multi-label learning," *Pattern Recognit.,* vol. 40, no. 7, pp. 2038-2048, 2007, doi: 10.1016/j.patcog.2006.12.019.

[56] L. Chen, "Predicting anatomical therapeutic chemical (ATC) classification of drugs by integrating chemical-chemical interactions and similarities," *PLoS ONE,* vol. 7, no. e35254, 2012.

[57] K. C. Chou, "Some remarks on predicting multi-label attributes in molecular biosystems," *Molecular Biosystems,* vol. 9, pp. 10922-1100, 2013.

[58] F. Gers, J. Schmidhuber, and F. Cummins, "Learning to Forget: Continual Prediction with LSTM," *Neural Computation,* vol. 12, no. 10, pp. 2451-2471, 2000, doi: 10.1162/089976600300015015.

[59] Y. Su, Y. Huang, and C.-C. J. Kuo, "On Extended Long Short-term Memory and Dependent Bidirectional Recurrent Neural Network," *Neurocomputing,* vol. 356, pp. 151–161, 2019.

[60] J. Chung, Ç. Gülçehre, K. Cho, and Y. Bengio, "Empirical Evaluation of Gated Recurrent Neural Networks on Sequence Modeling," *ArXiv,* vol. abs/1412.3555, 2014.

[61] N. Gruber and A. Jockisch, "Are GRU Cells More Specific and LSTM Cells More Sensitive in Motive Classification of Text?," *Frontiers in Artificial Intelligence,* vol. 3, no. 40, 2020, doi: 10.3389/frai.2020.00040.

[62] L. Jing *et al.*, "Gated Orthogonal Recurrent Units: On Learning to Forget," *Neural Computation,* vol. 31, no. 4, pp. 765-783, 2019, doi: doi.org/10.1162/neco_a_01174.

[63] K. Zhang, Z. Liu, and L. Zheng, "Short-Term Prediction of Passenger Demand in Multi-Zone Level: Temporal Convolutional Neural Network With Multi-Task Learning," *IEEE Transactions on Intelligent Transportation Systems,* vol. 21, no. 4, pp. 1480-1490, 2020, doi: 10.1109/TITS.2019.2909571.

[64] A. K. Jain, M. N. Murty, and P. J. Flynn, "Data clustering: a review," *ACM Comp Surv,* vol. 31, no. 3, pp. 264-323, 1999.

[65] S. Dubey, S. Chakraborty, S. K. Roy, S. Mukherjee, S. K. Singh, and B. Chaudhuri, "diffGrad: An Optimization Method for Convolutional Neural Networks," *IEEE Transactions on Neural Networks and Learning Systems,* vol. 31, no. 11, pp. 4500-4511, 2020.

[66] L. Nanni, G. Maguolo, and A. Lumini, "Exploiting Adam-like Optimization Algorithms to Improve the Performance of Convolutional Neural Networks," *ArXiv,* vol. abs/2103.14689, 2021.

[67] L. Nanni, A. Manfe, G. Maguolo, A. Lumini, and S. Brahnam, "High performing ensemble of convolutional neural networks for insect pest image detection," *ArXiv,* vol. abs/2108.12539, 2021.

[68] L. N. Smith, "Cyclical Learning Rates for Training Neural Networks," *2017 IEEE Winter Conference on Applications of Computer Vision (WACV),* pp. 464-472, 2017.

[69] M. Liu *et al.*, "Large-scale prediction of adverse drug reactions using chemical, biological, and phenotypic properties of drugs," *Journal of the American Medical Informatics Association : JAMIA,* vol. 19, pp. e28 - e35, 2012, doi: doi:10.1136/amiajnl-2011-000699.

[70] L. Yang, X.-Z. Wu, Y. Jiang, and Z. Zhou, "Multi-Label Learning with Deep Forest," *ArXiv,* vol. abs/1911.06557, 2020.

[71] F. K. Nakano, K. Pliakos, and C. Vens, "Deep tree-ensembles for multi-output prediction," *Pattern Recognit,* vol. 121, p. 108211, 2022/01/01/ 2022, doi: https://doi.org/10.1016/j.patcog.2021.108211.

[72] Jiang Wang, Yi Yang, Junhua Mao, Zhiheng Huang, Chang Huang, Wei Xu; "CNN-RNN: A Unified Framework for Multi-Label Image Classification" Proceedings of the IEEE Conference on Computer Vision and Pattern Recognition (CVPR), 2016, pp. 2285-2294

[73] Yazici, V. O., Gonzalez-Garcia, A., Ramisa, A., Twardowski, B., & Van De Weijer, J. (2020). Orderless recurrent models for multi-label classification. Proceedings of the IEEE Computer Society Conference on Computer Vision and Pattern Recognition, 13437-13446. https://doi.org/10.1109/CVPR42600.2020.01345